\pdfoutput=1
\documentclass[runningheads]{llncs}
\usepackage{graphicx}
\usepackage{amsmath,amssymb} 
\usepackage{color}
\usepackage{times}
\usepackage{epsfig}
\usepackage{amsmath}
\usepackage{amssymb}
\usepackage{mathptmx}
\usepackage{array}
\usepackage{multirow}
\usepackage{booktabs}
\usepackage{subcaption}
\usepackage{caption}

\usepackage[breaklinks=true]{hyperref}
\usepackage{breakcites}

\usepackage[normalem]{ulem}
\useunder{\uline}{\ul}{}

\long\def\ignorethis#1{}

\newlength\figwidth
\newlength\figmarginv
\newlength\figmarginh
\makeatother

\newlength\paramargin
\newlength\figmargin
\newlength\secmargin
\newlength\tabmargin
\newlength\figcapmargin

\newcommand\para[1]{\noindent{#1}}

\def\be {\begin{equation}}
\def\ee {\end{equation}}
\def\beas {\begin{eqnarray*}}
	\def\eeas {\end{eqnarray*}}
\def\bea {\begin{eqnarray}}
\def\eea {\end{eqnarray}}
\def\bes {\begin{equation*}}
\def\ees {\end{equation*}}
\def\ba {\begin{align}}
\def\ea {\end{align}}
\def\barr {\begin{array}}
	\def\earr {\end{array}}

\newcommand{\cB}{{\cal B}}
\newcommand{\cK}{{\cal K}}

\makeatletter
\usepackage{xspace}
\def\@onedot{\ifx\@let@token.\else.\null\fi\xspace}
\DeclareRobustCommand\onedot{\futurelet\@let@token\@onedot}

\newcommand{\figref}[1]{Figure~\ref{#1}} 
\newcommand{\equref}[1]{Eq\onedot~\eqref{#1}}
\newcommand{\secref}[1]{Section~\ref{#1}}
\newcommand{\tabref}[1]{Table~\ref{#1}}

\def\eg{\emph{e.g}\onedot} 
\def\ie{\emph{i.e}\onedot}

\def\etal{\emph{et al}\onedot}

\makeatother

\newcommand{\tb}[1]{\textbf{#1}}

\newcommand{\yt}[1]{{\color{cyan}{\bf Yuan-Ting: }#1}}

\begin{document}
\title{Unsupervised Video Object Segmentation using Motion Saliency-Guided Spatio-Temporal Propagation} 

\titlerunning{Unsupervised VOS using Motion Saliency-Guided Spatio-Temporal Propagation}
%
%

\author{Yuan-Ting Hu$^{1}$ \and
Jia-Bin Huang$^{2}$ \and
Alexander G. Schwing$^{1}$}

\authorrunning{Y.-T. Hu, J.-B. Huang, and A. G. Schwing}
%

\institute{
$^1$University of Illinois at Urbana-Champaign 
\hspace{20pt}
$^2$Virginia Tech \\
\hspace{10pt}
\email{\{ythu2,aschwing\}@illinois.edu}  \hspace{10pt}
\email{jbhuang@vt.edu}
}
\maketitle              
%
\begin{abstract}

Unsupervised video segmentation plays an important role in a wide variety of applications from object identification to compression. However, to date, fast motion, motion blur and occlusions pose significant challenges. To address these challenges for unsupervised video segmentation, we develop a novel saliency estimation technique as well as a novel neighborhood graph, based on optical flow and edge cues. Our approach leads to significantly better initial foreground-background estimates and their robust as well as accurate diffusion across time. We  evaluate our proposed  algorithm on the challenging DAVIS, SegTrack v2 and FBMS-59 datasets. Despite the usage of only a standard edge detector trained on 200 images, our method achieves state-of-the-art results outperforming deep learning based methods in the unsupervised setting. We even demonstrate competitive results comparable to deep learning based methods in the semi-supervised setting on the DAVIS dataset.

\end{abstract}

\vspace{\secmargin}
\section{Introduction}
\label{sec:intro}

Unsupervised foreground-background video object segmentation of complex scenes is a challenging problem which has many applications in areas such as object identification, security, and video compression. It is therefore not surprising that many efforts have been devoted to developing efficient techniques that are able to effectively separate foreground from background, even in complex videos.

\begin{figure}[t!]
\centering
\begin{minipage}{0.325\linewidth}\centering
\includegraphics[width=\linewidth]{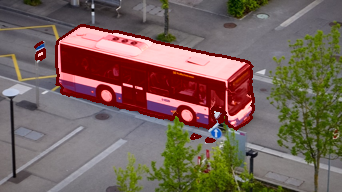} \\
\includegraphics[width=\linewidth]{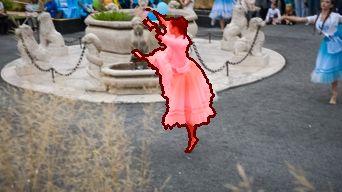} \\
\includegraphics[width=\linewidth]{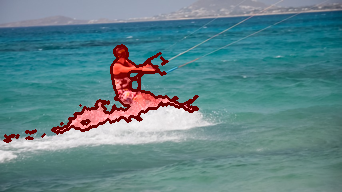}
\end{minipage}
\begin{minipage}{0.325\linewidth}\centering
\includegraphics[width=\linewidth]{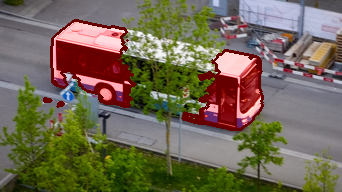} \\
\includegraphics[width=\linewidth]{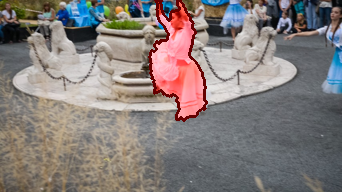} \\
\includegraphics[width=\linewidth]{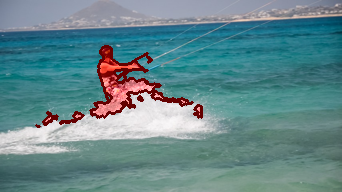}
\end{minipage}
\begin{minipage}{0.325\linewidth}\centering
\includegraphics[width=\linewidth]{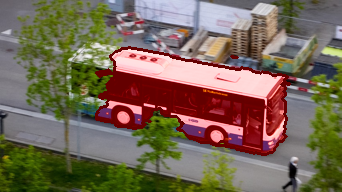} \\
\includegraphics[width=\linewidth]{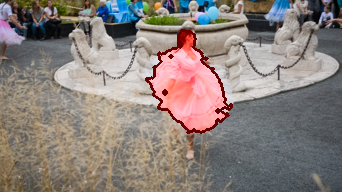} \\
\includegraphics[width=\linewidth]{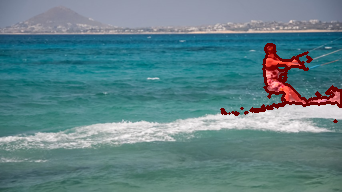}
\end{minipage}
\caption{\tb{Video object segmentation in challenging scenarios.} Given an input video, our algorithm produces accurate segmentation of the foreground object \emph{without any manual annotations}. Our method is capable of handling unconstrained videos that span a wide variety of situations including occlusion (\textsc{bus}),  non-ridge deformation (\textsc{dance-jump}), and dynamic background (\textsc{kite-surf}).
}
\label{fig:Teaser}
\end{figure}

In complex videos, cluttered backgrounds, deforming shapes, and fast motion are major challenges. In addition, in the unsupervised setting, algorithms have to automatically discover foreground regions in the video. To this end, classical video object segmentation techniques~\cite{BrutzerCVPR2011,CriminisiCVPR2006,ElgammalIEEE2002,HaymanICCV2003,IraniIJCV1994,RenPRL2003,IraniPAMI1998,TorrECCV1998,YuanPAMI2007} often assume rigid background motion models and incorporate a scene prior, two assumptions which are restrictive in practice. Trajectory based methods, such as~\cite{CosteiraICCV1995,ElhamifarCVPR2009,RaoCVPR2008,BroxECCV2010,fragkiadaki2012video}, require selection of clusters or a matrix rank, which may not be intuitive. Graphical model based approaches \cite{JainECCV14,BadrinarayananCVPR2010,GalassoICCV2013,TsaiBMVC2010,VijayanarasimhanECCV2012,TsaiCVPR2016} estimate the foreground regions using a probabilistic formulation. However, for computational efficiency, the constructed graph usually contains only local connections, both spatially and temporally, reducing the ability to consider long-term spatial and temporal coherence patterns. To address this concern, diffusion based methods~\cite{Lovász93randomwalks}, \eg,~\cite{FaktorBMVC14,wang2012probabilistic}, propagate an initial foreground-background estimate more globally. While promising results are shown, diffusion based formulations rely heavily on the initialization as well as an accurate neighborhood graph encoding the semantic distance between pixels or superpixels.


Therefore, in this paper, we develop (1) a new initialization technique and (2) a more robust neighborhood graph. Our initialization technique is based on the intuition that the optical flow on the boundary of an image differs significantly from the moving direction of the object of interest. Our robust neighborhood graph is built upon accurate edge detection and flow cues. 

%
We highlight the performance of our proposed approach in~\figref{fig:Teaser} using three challenging video sequences. Note the fine details that our approach is able to segment despite the fact that our method is unsupervised. Due to accurate initial estimates and a more consistent neighborhood graph, we found our method to be robust to different parameter choices.  Quantitatively, our initialization technique and neighborhood graph result in significant improvements for unsupervised foreground-background video segmentation when compared to the current state-of-the-art. On the recently released DAVIS dataset~\cite{PerazziCVPR16}, our unsupervised non-deep learning based segmentation technique outperforms current state-of-the-art methods by more than 1.3\% in the unsupervised setting.
Our method also achieves competitive performance compared with deep net based techniques in the semi-supervised setting.



\section{Related Work}
\label{sec:related}

The past decade has seen the rapid development in video object segmentation \cite{TsaiBMVC2010,LiICCV13,NagarajaICCV15,PriceICCV09,LezamaCVPR11,LeeICCV11,PapazoglouICCV13,XiaoCVPR16,GrundmannCVPR2010,TsaiCVPR2016,jain2017fusionseg,HuNIPS2017,HuECCV2018b}. 
Given different degrees of human interaction, these methods model inter- and  intra-frame relationship of the pixels or superpixels to determine the foreground-background labeling of the observed scene. Subsequently, we classify the literature into four areas based on the degree of human involvement and discuss the relationship between video object  and video motion segmentation.

\noindent{\bf Unsupervised video object segmentation:}
Fully automatic approaches for video object segmentation have been explored recently~\cite{ChengCVPR12,LeeICCV11,ZhangCVPR13,PapazoglouICCV13,OchsPAMI2014,FaktorBMVC14,XiaoCVPR16,KohCVPR17}, and no manual annotation is required in this setting. Unsupervised foreground segmentation discovery can be achieved by motion analysis~\cite{PapazoglouICCV13,FaktorBMVC14}, trajectory clustering~\cite{OchsPAMI2014}, or object proposal ranking~\cite{LeeICCV11,XiaoCVPR16}. 
Our approach computes motion saliency in a given video based on boundary similarity of motion cues. In contrast, Faktor and Irani~\cite{FaktorBMVC14} find motion salient regions by extracting dominant motion. Subsequently they obtain the saliency scores by computing the motion difference with respect to the detected dominant motion. Papazoglou and Ferrari~\cite{PapazoglouICCV13} identify  salient regions by finding the motion boundary based on optical flow and computing  inside-outside maps to detect the object of interest. 

Recently, deep learning based methods~\cite{jain2017fusionseg,TokmakovCVPR2017,Tokmakov17_2} were also used to address unsupervised video segmentation. Although these methods do not require the ground truth of the first frame of the video (unsupervised as opposed to semi-supervised), they need a sufficient amount of labeled data to train the models.
In contrast, our approach works effectively in the unsupervised setting and does not require training data beyond the one used to obtain an accurate edge detector.

\noindent{\bf Tracking-based video object segmentation:}
In this setting, the user annotation is reduced to only one mask for the first frame of the video~\cite{BrendelICCV09,GrundmannCVPR2010,TsaiBMVC2010,JainECCV14,TsaiCVPR2016,MaerkiCVPR16,PerazziCVPR2017}. These approaches track the foreground object and propagate the segmentation results to successive frames by incorporating cues such as motion~\cite{TsaiBMVC2010,TsaiCVPR2016} and supervoxel consistency~\cite{JainECCV14}. Again, our approach differs in that we don't consider any human labels.

\noindent{\bf Interactive video object segmentation:}
Interactive video object segmentation allows users to annotate the foreground segments in key frames to generate impressive results by propagating the user-specified masks across the entire video~\cite{PriceICCV09,FanTOG15,NagarajaICCV15,LiTOG16,JainECCV14}. Price~\etal~\cite{PriceICCV09} further combine multiple features, of which the weights are automatically selected and learned from user inputs. Fan~\etal~\cite{FanTOG15} tackle interactive segmentation by enabling bi-directional propagation of the masks between non-successive frames. Our approach differs in that the proposed method does not require any human interaction.

\noindent{\bf Video motion segmentation:}
Video motion segmentation~\cite{BroxECCV2010} aims to segment a video based on motion cues, while video object segmentation aims at segmenting the foreground based on objects. The objective function differs: for motion segmentation, clustering based methods~\cite{BroxECCV2010,OchsPAMI2014,KeuperICCV15,LezamaCVPR11} are predominant and group point trajectories. In contrast, for video object segmentation, a binary labeling formulation is typically applied as we show next by describing our approach. 

\begin{figure*}[t]
\centering
\includegraphics[height=1.01in]{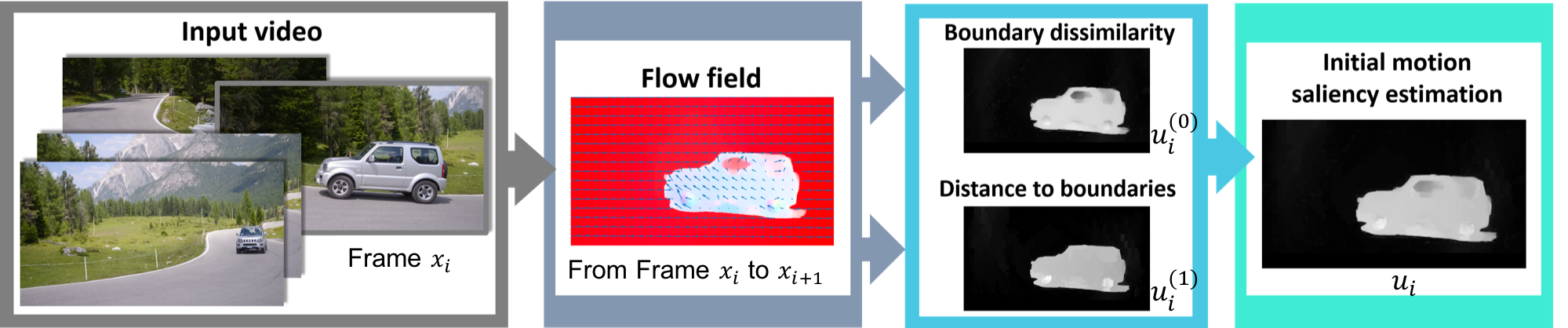}\\
\caption{{\tb{Motion saliency estimation}}. Given an input video, we compute the flow field for each frame. We detect the saliency score based on the flow vector by calculating a boundary dissimilarity map $u^{(0)}$ and a distance map $u^{(1)}$ indicating the distance of each pixel to the boundaries. We use minimum barrier distance to measure the distance. The motion saliency estimation is computed by averaging the boundary dissimilarity map and the distance map.}
\label{fig:init}
\end{figure*}
\section{Unsupervised Video Object Segmentation}
\label{sec:method}

The two most important ingredients for unsupervised video object segmentation are the initial saliency estimate as well as a good assessment of the neighborhood relation of pixels or superpixels. For initial saliency prediction in unsupervised video object segmentation we describe a novel method comparing the motion  at a pixel to the boundary motion. Intuitively, boundary pixels largely correspond to background and pixels with a similar motion are likely background too. To construct a meaningful neighborhood relation between pixels we assess flow and appearance cues. We provide details for both contributions after describing an overview of our unsupervised video object segmentation approach. 

\para{\bf Method overview:} 
Our method uses a diffusion mechanism for unsupervised video segmentation. Hence, the approach distributes an initial foreground saliency estimate over the $F$ frames $x_i$, $i\in\{1, \ldots, F\}$, of a video $x = (x_1, \ldots, x_F)$. To this end, we partition each frame into a set of nodes using superpixels, and estimate and encode their semantic relationship within and across frames using a global neighborhood graph. Specifically, we represent the global neighborhood graph by a weighted row-stochastic adjacency matrix $G \in \mathbb{R}^{N\times N}$, where $N$ is the total number of nodes in the video. Diffusion of the initial foreground saliency estimates $v^0 \in \mathbb{R}^{N}$ for each node is performed by repeated matrix multiplication of the current node estimate with the adjacency matrix $G$, \ie,   for the $t$-th diffusion step $v^t = Gv^{t-1}$.

With the adjacency matrix $G$ and initialization $v^0$ being the only inputs to the algorithm, it is obvious that they 
are of crucial importance for diffusion based unsupervised video segmentation. We focus on both points in the following and develop first a new saliency estimation of $v^0$ before discussing  construction of the neighborhood graph $G$.

\subsection{Saliency estimation}
\label{sec:init}
For unsupervised video object segmentation, we propose to estimate the motion saliency by leveraging a boundary condition. 
Since we are dealing with video, motion is one of the most important cues for identifying moving foreground objects. In general, the motion of the foreground object differs from background motion. But importantly, the background region is often connected to the boundary of the image. While the latter assumption is commonly employed for \emph{image saliency} detection, it has not been exploited for \emph{motion saliency} estimation. 
To obtain the initial saliency estimate $v^0$ defined over superpixels, we average the pixelwise motion saliency results $u$ over the spatial support of each superpixel. We subsequently describe our developed procedure for foreground saliency estimation, taking advantage of the boundary condition. The proposed motion saliency detection is summarized in~\figref{fig:init}.

Conventional motion saliency estimation techniques for video object segmentation are based on either background subtraction~\cite{BrutzerCVPR2011}, trajectory clustering~\cite{BroxECCV2010}, or motion separation~\cite{FaktorBMVC14}. Background subtraction techniques typically assume a static camera, which is not applicable for complex videos. Trajectory clustering groups points with similar trajectories, which is sensitive to non-rigid transformation.
Motion separation detects background by finding the dominant motion and subsequently calculates the difference in magnitude and/or orientation between the motion at each pixel, and the dominant motion. The larger the difference, the more likely the pixel to be foreground.
Again, complex motion poses challenges, making it hard to separate foreground from background. 

In contrast, we propose to use the boundary condition that is commonly used for \emph{image saliency} detection \cite{WeiECCV2012,TuCVPR16} to support \emph{motion saliency} estimation for unsupervised video segmentation. Our approach is based on the intuition that the background region is connected to image boundaries in some way. Therefore we calculate a distance metric for every pixel to the boundary. 
Compared to the aforementioned techniques, we will show that our method can better deal with complex, non-rigid motion. 

We use $u$ to denote the foreground motion saliency of the video. 
Moreover, $u_{i}$ and $u_{i}(p_i)$ denote the foreground saliency for frame $i$ and for pixel $p_i$ in frame $i$ respectively. 
To compute the motion saliency estimate, we treat every frame $x_i$, $i\in\{1, \ldots, F\}$ independently. Given a frame $x_i$, let $x_{i}(p_i)$ refer to the intensity values of pixel $p_i$, and let $f_{i}(p_i)\in\mathbb{R}^2$ denote the optical flow vector measuring the motion of the object illustrated at pixel $p_i$ between frame $i$ and frame $i+1$. In addition, let $\cB_i$ denote the set of boundary pixels of frame $i$. 

We compute the foreground motion saliency $u_{i}$ of frame $i$ based on two terms $u_{i}^{(0)}$ and $u_{i}^{(1)}$, each of which measures a distance between any pixel $p_i$ of the $i$-th frame and the boundary $\cB_i$. For the first distance $u_{i}^{(0)}$, we compute the smallest flow direction difference observed between a pixel $p_i$ and common flow directions on the boundary. For the second distance $u_{i}^{(1)}$, we measure the smallest barrier distance between pixel $p_i$ and boundary pixels. Both of the terms capture the similarity between the motion at pixel $p_i$ and the background motion. Subsequently, we explain both terms in greater detail.


\para{\bf Computing flow direction difference:} More formally, to compute $u_{i}^{(0)}(p_i)$, the flow direction difference between pixel $p_i$ in frame $i$ and common flow directions on the boundary $\cB_i$ of frame $i$, we first cluster the boundary flow directions into a set of $K$ clusters $k\in\{1, \ldots, K\}$ using k-means. We subsume the cluster centers in the set
\be
\cK_i \!=\! \left\{\!\mu_{i,k}\! : \!\mu_{i,k} \!=\! \arg\min_{\hat \mu_{i,k}}\!\min_{r\in\{0,1\}^{|\cB_i|K}}\! \frac{1}{2}\!\!\sum_{p_i\in\cB_i,k} \!\!\!\!\!\!r_{p_i,k}\|f_{i}(p_i) - \hat\mu_{i,k}\|_2^2\!\!\right\}\!.
\ee
Hereby, $r_{p_i,k}\in\{0,1\}$ is an indicator variable which assigns pixel $p_i$ to cluster $k$, and $r$ is the concatenation of all those indicator variables. We update $\cK_i$ to only contain centers with more than $1/6$ of the boundary pixels assigned. Given those cluster centers, we then obtain a first distance measure capturing the difference of flow between pixel $p_i$ in frame $i$ and the major flow directions observed at the boundary of frame $i$ via
\be
u_{i}^{(0)}(p_i) = \min_{\mu_{i,k}\in\cK_i} \|f_{i}(p_i) - \mu_{i,k}\|_2^2.
\ee

\para{\bf Computing smallest barrier distance:} When computing the smallest barrier distance $D_{bd,i}$ between pixel $p_i$ in frame $i$ and boundary pixels, \ie, to obtain
\be
u_{i}^{(1)}(p_i) = \min_{s \in \cB_i} D_{bd,i}(p_i, s),
\ee
we use the following barrier distance:
\be
D_{bd,i}(p_i,s) = \max_{e \in \Pi_{i,p_i,s}} w_i(e) - \min_{e \in \Pi_{i,p_i,s}} w_i(e).
\label{eq:BarrDist}
\ee
Hereby, $\Pi_{i,p_i,s}$ denotes the path, \ie, a set of edges connecting pixel $p_i$ to boundary pixel $s\in\cB_i$, obtained by computing a minimum spanning tree on frame $i$. The edge weights $w_i(e)$, which are used to compute both the minimum spanning tree as well as the barrier distance given in \equref{eq:BarrDist}, are obtained as the maximum flow direction difference between two neighboring pixels, \ie, $w_i(e) = \max\left\{f_{i}(p_i) - f_{i}(q_i)\right\}\in\mathbb{R}$ where the $\max$ is taken across the two components of $f_{i}(p_i) - f_{i}(q_i) \in \mathbb{R}^2$. Note that $e = (p_i,q_i)$ refers to an edge connecting the two pixels $p_i$ and $q_i$. To compute the minimum spanning tree we use the classical 4-connected neighborhood. Intuitively, we compute the barrier distance between 2 points as the difference between the maximum edge weight and minimum edge weight on the path of the minimum spanning tree between the 2 points. We then compute the smallest barrier distance of a point as the minimum of the barrier distances between the point and any point on the boundary. 

\para{\bf Computing foreground motion saliency:} We obtain the pixelwise foreground motion saliency $u_{i}$ of frame $i$ when adding the two distance metrics $u_{i}^{(0)}$ and $u_{i}^{(1)}$ after having normalized each of them to a range of $[0,1]$ by subtracting the minimum entry in $u_{i}^{(\cdot)}$ and dividing by the difference between the maximum and minimum entry. Examples for $u_{i}^{(0)}$, $u_{i}^{(1)}$ and the combined motion saliency are visualized in \figref{fig:init}. 

We found the proposed changes to result in significant improvements for saliency estimation of video data. We present a careful assessment in \secref{sec:exp}.

\subsection{Neighborhood construction}
\label{sec:graph}
\begin{figure*}[t]
\centering
\vspace{\figmargin}
\includegraphics[height=1.01in]{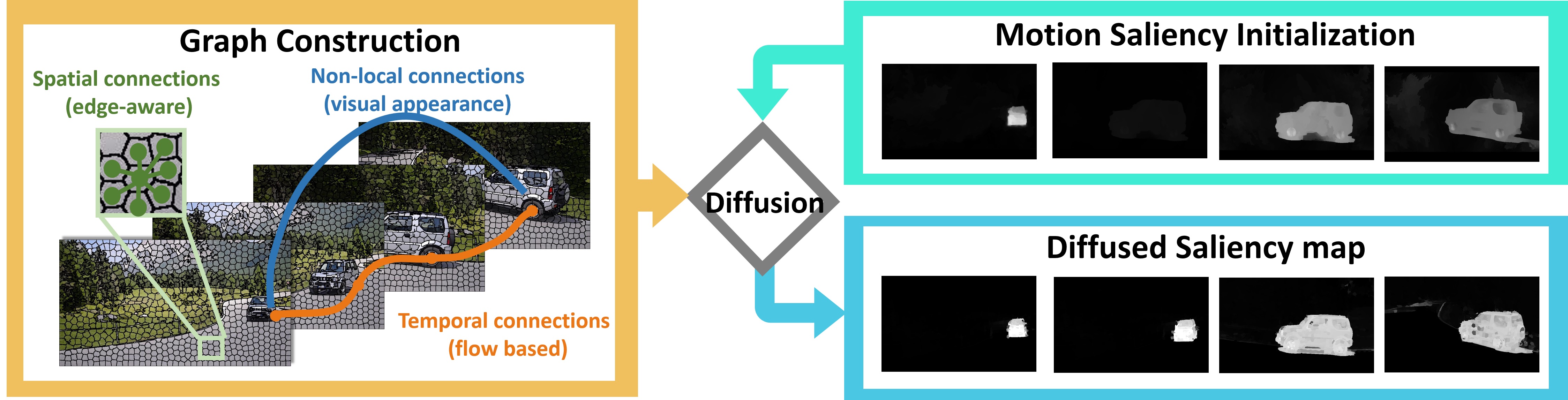}\\
\vspace{\figcapmargin}
\caption{{\tb{Graph construnction}}. In our method, we construct a graph for diffusing the initial motion saliency estimation. Our graph contains 1) edge-aware spatial connections (intra-frame connections), 2) flow-based temporal connections (inter-frame connections and 3) non-local long range connections. We show the initial motion saliency and the diffused saliency map using the constructed graph. We found  these three types of connections to help propagate the initial saliency estimation  effectively.}
\label{fig:graph}
\vspace{\figmargin}
\end{figure*}
The second important term for diffusion based video segmentation beyond initial estimates is the neighborhood graph $G$. Classical techniques construct the adjacency matrix using 
local information, such as connecting a node with its spatial and temporal neighbors, 
and non-local connections. 
These methods establish a connection between two nodes as long as their visual appearance is similar. 

In contrast, we compute the neighborhood graph, \ie, the adjacency matrix for graph diffusion, $G = T\times E \times V$ as the product of three components, based on inter-frame information $T$, intra-frame signals $E$, and long-range components $V$, as shown in~\figref{fig:graph}, and use a variety of cues for robustness. We formally discuss each of the components in the following. 

\para{\bf Inter-frame temporal information} is extracted from optical flow cues. We connect superpixels between adjacent frames following flow vectors while checking the forward/backward consistency in order to prevent inaccurate flow estimation at motion boundaries.

More formally, to compute the flow adjacency matrix $T$, consider two successive video frames $x_i$ and $x_{i+1}$ each containing pixel $p_i$ and $p_{i+1}$, respectively. We compute a forward flow field $f_i(p_i)$ and a backward flow field $b_{i+1}(p_{i+1})$ densely for every pixel $p$ using~\cite{hu2016efficient}. Using those flow fields, we define the forward confidence score $c_i^F(p_i)$ at pixel $p_i$ of frame $x_i$ via
\be
c_i^F(p_i) = \exp\left(\frac{-\|-f_i(p_i)-b_{i+1}(p_i+f_i(p_i))\|_2^2}{\sigma_2}\right),
\label{eq:Confidence}
\ee
and the backward confidence score $c_i^B(p_i)$ at pixel $p_i$ of frame $x_i$ via
\be
c_i^B(p_i) = \exp\left(\frac{-\|-b_i(p_i)-f_{i-1}(p_i+b_i(p_i))\|_2^2}{\sigma_2}\right),
\label{eq:Confidence_B}
\ee
where $\sigma_2$ is a hyper-parameter. Intuitively, this confidence score measures the distance between the pixel $p_i$ and the result obtained after following the flow field into frame $x_{i+1}$ via $p_i+f_i(p_i)$ and back into frame $x_i$ via $p_i+f_i(p_i)+b_{i+1}(p_i+f_i(p_i))$. Taking the difference between pixel $p_i$ and the obtained reprojection results in the term given in \equref{eq:Confidence} and \equref{eq:Confidence_B}. 
We use the confidence scores to compute the connection strength between two superpixels $s_{i,k}$ and $s_{i+1,m}$ in frame $i$ and $i+1$ via 
\be \small
T(s_{i,k},s_{i+1,m}) \!=\!\!\!
\sum_{p\in s_{i,k}} \!\!\frac{\delta(p + f_i(p)\in s_{i+1,m})c_i^F(p)}{|s_{i,k}| + |s_{i+1,m}|}+ \!\!\!\!\!\sum_{p^\prime\in s_{i+1,m}} \!\!\!\!\!\frac{\delta(p^\prime + b_{i+1}(p^\prime)\in s_{i,k})c_{i+1}^B(p^\prime)}{|s_{i,k}| + |s_{i+1,m}|}.
\ee
Hereby $\delta(\cdot)$ denotes the indicator function and $|s_{i,k}|$ and $|s_{i+1,m}|$ 
represent the number of pixels in $s_{i,k}$ and $s_{i+1,m}$, respectively. Intuitively, the first term compares the strength of the connections that start in superpixel $s_{i,k}$ and end up in superpixel $s_{i+1,m}$ with the total amount of strength originating from both $s_{i,k}$ and $s_{i+1,m}$. Similarly for the second term. 

\para{\bf Intra-frame spatial information} prevents diffusion across visual edges within a frame, while allowing information to be propagated between adjacent superpixels in the same frame if they aren't separated by a strong edge. 

More formally,
to find the edge aware spatial connections $E$, we first detect the edge responses frame-by-frame using the training based method discussed in~\cite{DollarICCV13edges}. 
Given edge responses, we calculate the confidence scores $A(s)$ for all superpixel $s$ by summing over the decay function, \ie,
\be
A(s) =\frac{1}{|s|} \sum_{p\in s}\frac{1}{1+exp(\sigma_w\cdot(G(p)-\epsilon))}.
\label{eq:EdgeConnection}
\ee
Hereby, $G(p)\in[0,1]$ is the edge response at pixel $p$. 
$\sigma_w$ and $\epsilon$ are hyper-parameters, which we fix at $\sigma_w = 50$ and $\epsilon=0.05$ for all our experiments. 

We calculate the edge-aware adjacency matrix $E$ by exploiting the above edge information. Specifically, 
\be
E(s_{i,k}, s_{i,m}) = \frac{1}{2} \left(A(s_{i,k}) + A(s_{i,m})\right),
\ee
if $s_{i,k}$ is spatially close to $s_{i,m}$, \ie, if the distance between the centers of the two superpixels is less than 1.5 times the square root of the size of the superpixel.

\para{\bf Long range connections} based on visual similarity allow propagating information between superpixels that are far away either temporally or spatially as long as the two are visually similar. These long-range connections enable the information to propagate more efficiently through the neighborhood graph.

More formally,
to compute the visual similarity matrix $V$, we find those superpixels that are most closely related to a superpixel $s_{i,m}$. To this end, we first perform a $k$ nearest neighbor search. More specifically, for each superpixel $s_{i,m}$ we find its $k$ nearest neighbors that are within a range of $r$ frames temporally. 
To compute the distance between two superpixels we use the Euclidean distance in the feature space. 

We compute features $f(s)$ of a superpixel $s$ by concatenating the LAB and RGB histograms computed over the pixels within a superpixel. We also include the HOG feature, and the $x$ and $y$ coordinate of the center of the superpixel.

Let the $k$ nearest neighbors of the superpixel $s_{i,m}$ be referred to via $N(s_{i,m})$.
The visual similarity matrix is then defined via
\be
\small
V(s_{i,m},s) = \exp\left(\frac{-\|f(s_{i,m}) - f(s)\|_2^2}{\sigma}\right)\quad\forall s\in N(s_{i,m}),
\label{eq:NonlocalConnection}
\ee
where $\sigma$ is a hyper-parameter and $f(s)$ denotes the feature representation of the superpixel $s$. Note that we use the same features to find $k$ nearest neighbors and to compute the visual similarity matrix $V$. 
In this work, we refrain from using deep net based information even though we could easily augment our technique with more features.

To address the computational complexity, we use an approximate $k$ nearest neighbor search. Specifically, we use the fast implementation of ANN search utilizing the randomized k-d forest provided in~\cite{MujaFLANN09}.


\section{Experiments}
\label{sec:exp}

In the following, we present the implementation details, describe the datasets and metrics used for evaluation, followed by ablation study highlighting the influences of the proposed design choices and comparisons with the state-of-the-art.

\subsection{Implementation details}
For the proposed saliency estimation algorithm, we set the number of clusters $K=3$ for modeling the background. 
For neighborhood graph construction described in Section~\ref{sec:graph}, we  found $k=40, r=15, \sigma=0.1, \sigma_2=2^{-6}, \sigma_w=50$ to work well across datasets.  The number of diffusion iterations is set to 25.
In the supplementary material, we show that the performance of our method is reasonably robust to parameter choices. 
%

The average running time of our approach on the DAVIS dataset, including the graph construction and diffusion is about $8.5$ seconds per frame when using a single PC with Intel i$7$-$4770$ CPU and $32$ GB memory. 
%
%
Extracting superpixels and feature descriptors takes about $1.5$ and $0.8$ seconds per frame, respectively. 
We use the implementation by~\cite{hu2016efficient,RevaudCVPR2015} for computing optical flow, which takes about $10.7$ seconds per frame, including both forward flow and backward flow. 

\subsection{Datasets}
 We extensively compare our proposed technique to a series of baselines using the DAVIS dataset~\cite{PerazziCVPR16} (50 video sequences), the SegTrack v2 dataset~\cite{LiICCV13} (14 video sequences), and the FBMS-59 dataset~\cite{OchsPAMI2014} (22 video sequences in the test set).
These datasets are challenging as they contain nonrigid deformation, drastic illumination changes, cluttered background, rapid object motion, and occlusion. 
All three datasets provide pixel-level ground-truth annotations for each frame. 

\subsection{Evaluation metrics}

\para{\bf Intersection over union ($\mathcal{J}$):} The intersection over union (IoU) metric, also called the Jaccard index, computes the average over the dataset. The IoU metric has been widely used for evaluating the quality of the segmentation. 

\para{\bf Contour accuracy ($\mathcal{F}$)~\cite{PerazziCVPR16}:} To assess the segmentation quality, we compute the contour accuracy as $\mathcal{F}=\frac{2PR}{P+R}$, where $P$ and $R$ are the matching precision and recall of the two sets of points on the contours of the ground truth segment and the output segment, calculated via a bipartite graph matching.

\para{\bf Temporal stability ($\mathcal{T}$)~\cite{PerazziCVPR16}:} The temporal stability is measured by computing the distance between the shape context descriptors~\cite{Belongie02PAMI} describing the shape of the boundary of the segmentations between  two successive frames. Intuitively, the metric indicates the degree of deformation required to transform the segmentation mask from one frame to its adjacent frames.

Subsequently we first present an ablation study where we assess the contributions of our technique. 
Afterwards we perform a quantitative evaluation where we compare the accuracy of our approach to baseline video segmentation approaches. Finally we present qualitative results to illustrate the success and failure cases of our method.

\begin{table}[t]
	\centering
{\small

	\caption{Contribution of different components of our algorithm evaluated on the DAVIS dataset. Our algorithm with inter-frame, intra-frame connections, long range connections, and focused diffusion (denoted as FDiff) enabled performs best and achieves an IoU of {\bf {77.56}}\%. 
	}
	\label{tab:Ablation}
	\tabcolsep=3pt
	\begin{tabular}{ccc|c|c}
		\toprule
		\multicolumn{3}{c|}{Connections}& \multirow{2}{*}{FDiff} & \multirow{2}{*}{IoU ($\%$)}\\
		\cline{1-3}
		Inter-frame&Intra-frame&Long range&  &\\
		\hline
		-&-&-&-&{57.52}\\
		\checkmark&-&-&-&{62.75}\\
		-&\checkmark&-&-&{62.13}\\
		-&-&\checkmark&-&{72.38}\\
		\checkmark&\checkmark&-&-&{65.01}\\
		\checkmark&-&\checkmark&-&{72.70}\\
		-&\checkmark&\checkmark&-&{74.13}\\
		\checkmark&\checkmark&\checkmark&-&74.34\\
		\checkmark&\checkmark&\checkmark&\checkmark&{\bf 77.56}\\
		\bottomrule 
	\end{tabular}	}
\end{table}

\subsection{Ablation study}
We assess the resulting performance of the individual components of our adjacency defined neighborhood in~\tabref{tab:Ablation}. The performance in IoU of the motion saliency estimation in our approach (with all the connections disabled) is $57.52\%$. We analyze the effect of the three main components in the adjacency graph: (1) inter-frame flow based temporal connections $T$, (2) intra-frame edge based spatial connections $E$ and (3) long range connections $V$.

The improvements reported for saliency estimation and neighborhood construction motivate their use for unsupervised video segmentation. Besides, we  apply a second round of `focused diffusion,' restricted to the region which focuses primarily on the foreground object, to improve the results. The effects of the focused diffusion (denoted `FDiff') can be found in~\tabref{tab:Ablation} as well, showing significant improvements.

In~\tabref{tab:Ablation}, the checkmark `\checkmark' indicates the {\emph{enabled}} components. We observe consistent improvements when including additional components, which improve the robustness of the proposed method. 

\begin{table*}[t]
	\centering
		\caption{The quantitative evaluation on the DAVIS dataset~\cite{PerazziCVPR16}. Evaluation metrics are the IoU measurement $\mathcal{J}$, boundary precision $\mathcal{F}$, and time stability $\mathcal{T}$. Following~\cite{PerazziCVPR16}, we also report the recall and the decay of performance over time for $\mathcal{J}$ and $\mathcal{F}$ measurements. 
		}
	\label{tab:IoU}
\footnotesize
\setlength\tabcolsep{-0.3pt}
\resizebox{\linewidth}{!}{
\begin{tabular}{@{\extracolsep{4pt}}cc|cccccccccccccccccc}
\toprule		
&&\multicolumn{9}{c}{Semi-supervised}&\multicolumn{8}{c}{Unsupervised}\\
\cline{3-12} \cline{13-20}
&&SEA&HVS&JMP&FCP&BVS&OFL&CTN&VPN&MSK&\bf OURS-S&NLC&MSG&KEY&FST&FSG&LMP&ARP&\bf OURS-U\\


\multicolumn{2}{c|}{{Deep features}}&-&-&-&-&-&\checkmark&\checkmark&\checkmark&\checkmark&-&-&-&-&-&\checkmark&\checkmark&-&-\\
\midrule


& Mean $\mathcal{M} \uparrow$     &   \ 0.556 &   \ 0.596 &   \ 0.607 &   \ 0.631 &   \ 0.665 &   \ 0.711 &   \ 0.755 &   \ 0.750 &{\ul\ 0.803 }& \bf   \ 0.810 &   \ 0.641 &   \ 0.543 &   \ 0.569 &   \ 0.575 &   \ 0.716 &   \ 0.697 &  {\ul\ \ 0.763} &\bf\ 0.776\\
$\mathcal{J}$ & Recall $\mathcal{O} \uparrow$   &   \ 0.606 &   \ 0.698 &   \ 0.693 &   \ 0.778 &   \ 0.764 &   \ 0.800 &   \ 0.890 &   \ 0.901 &{\ul\ 0.935} & \bf  \ 0.946 &   \ 0.731 &   \ 0.636 &   \ 0.671 &   \ 0.652 &   \ 0.877 &   \ 0.829 &   \bf\ 0.892 & {\ul\ 0.886} \\
& Decay $\mathcal{D} \downarrow$  &   \ 0.355 &   \ 0.197 &   \ 0.372 & \bf  \ 0.031 &   \ 0.260 &   \ 0.227 &   \ 0.144 &  {\ul \ 0.093} & \ 0.089 &   \ 0.102 &   \ 0.086 &   {\ul\ 0.028} &   \ 0.075 &   \ 0.044 &\bf\ 0.017 &   \ 0.056 &   \ 0.036 &   \ 0.044 \\
\midrule
& Mean $\mathcal{M} \uparrow$     &   \ 0.533 &   \ 0.576 &   \ 0.586 &   \ 0.546 &   \ 0.656 &   \ 0.679 &   \ 0.714 &   \ 0.724 &  {\ul \ 0.758} &\bf\ 0.783 &   \ 0.593 &   \ 0.525 &   \ 0.503 &   \ 0.536 &   \ 0.658 &   \ 0.663 &  {\ul\ \ 0.711} &\bf\ 0.750 \\
 $\mathcal{F}$ & Recall $\mathcal{O} \uparrow$   &   \ 0.559 &   \ 0.712 &   \ 0.656 &   \ 0.604 &   \ 0.774 &   \ 0.780 &   \ 0.848 &   \ 0.842 &  {\ul \ 0.882} &\bf\ 0.928 &   \ 0.658 &   \ 0.613 &   \ 0.534 &   \ 0.579 &   \ 0.790 &   \ 0.783 &  {\ul\ \ 0.828} &\bf\ 0.869 \\
& Decay $\mathcal{D} \downarrow$  &   \ 0.339 &   \ 0.202 &   \ 0.373 &\bf\ 0.039 &   \ 0.236 &   \ 0.240 &   \ 0.140 &   \ 0.136 &   \ {\ul\ 0.095} &   \ 0.115 &   \ 0.086 &   \ 0.057 &   \ 0.079 &   \ 0.065 & {\ul\ \ 0.043} &   \ 0.067 &   \ 0.073 &  \bf \ 0.042 \\
\midrule
$\mathcal{T}$ & Mean $\mathcal{M} \downarrow$   & {\ul\  \ 0.137} &   \ 0.296 &\bf\ 0.131 &   \ 0.285 &   \ 0.316 &   \ 0.239 &   \ 0.198 &   \ 0.300 &   \ 0.189 &   \ 0.212 &   \ 0.356 &   \ 0.250 &\bf\ 0.190 &   \ 0.276 &   \ 0.286 &   \ 0.689 &   \ 0.352 &   {\ul \ 0.243} \\
\bottomrule
	\end{tabular}
}

\end{table*}

\begin{table*}[t]
	\centering
	\caption{The attribute-based aggregate performance comparing  unsupervised methods on the DAVIS dataset~\cite{PerazziCVPR16}. We calculate the average IoU of all sequences with the specific attribute: appearance change (AC), dynamic background (DB), fast motion (FM), motion blur (MB), and occlusion (OCC). The right column with small font indicates the performance change for the method on the remaining sequences if the sequences possessing the corresponding attribute are not taken into account.}
	\label{tab:attr}
	\setlength\tabcolsep{5pt}
	\setlength\extrarowheight{-1.5pt}
	\renewcommand\arraystretch{0.9}
	\resizebox{\linewidth}{!}{
	\begin{tabular}{lcccccccc}	\toprule	Attribute&NLC~\cite{FaktorBMVC14}&MSG~\cite{BroxECCV2010}&KEY~\cite{LeeICCV11}&FST~\cite{PapazoglouICCV13}&FSG~\cite{jain2017fusionseg}&LMP~\cite{TokmakovCVPR2017}&ARP~\cite{KohCVPR17}&\bf OURS-U\\
\midrule

AC	&    0.54 \emph{\scriptsize{+0.13}}	&    0.48 \emph{\scriptsize{+0.08}}	&    0.42 \emph{\scriptsize{+0.19}}	&    0.55 \emph{\scriptsize{+0.04}}	&\bf 0.73 \emph{\scriptsize{-0.02}}	&    0.67 \emph{\scriptsize{+0.03}}	&    {\ul\ 0.73 \emph{\scriptsize{+0.04}}}	&    0.72 \emph{\scriptsize{+0.07}}	\\
DB	&    0.53 \emph{\scriptsize{+0.15}}	&    0.43 \emph{\scriptsize{+0.15}}	&    0.52 \emph{\scriptsize{+0.07}}	&    0.53 \emph{\scriptsize{+0.06}}	&    {\ul\ 0.67 \emph{\scriptsize{+0.05}}}	&    0.57 \emph{\scriptsize{+0.16}}	&\bf 0.70 \emph{\scriptsize{+0.08}}	&    0.66 \emph{\scriptsize{+0.15}}	\\
FM	&    0.64 \emph{\scriptsize{+0.00}}	&    0.46 \emph{\scriptsize{+0.14}}	&    0.50 \emph{\scriptsize{+0.12}}	&    0.50 \emph{\scriptsize{+0.12}}	&    0.69 \emph{\scriptsize{+0.04}}	&    0.67 \emph{\scriptsize{+0.05}}	&    {\ul\ 0.73 \emph{\scriptsize{+0.05}}}	&\bf 0.75 \emph{\scriptsize{+0.04}}	\\
MB	&    0.61 \emph{\scriptsize{+0.04}}	&    0.35 \emph{\scriptsize{+0.29}}	&    0.51 \emph{\scriptsize{+0.08}}	&    0.48 \emph{\scriptsize{+0.14}}	&    0.65 \emph{\scriptsize{+0.10}}	&    0.64 \emph{\scriptsize{+0.08}}	&   {\ul\ 0.69 \emph{\scriptsize{+0.11}}}	&\bf 0.74 \emph{\scriptsize{+0.06}}	\\
OCC	&    0.70 \emph{\scriptsize{-0.09}}	&    0.48 \emph{\scriptsize{+0.10}}	&    0.52 \emph{\scriptsize{+0.08}}	&    0.53 \emph{\scriptsize{+0.07}}	&    0.65 \emph{\scriptsize{+0.10}}	&    0.70 \emph{\scriptsize{-0.01}}	&    {\ul\ 0.71 \emph{\scriptsize{+0.08}}}	&\bf 0.81 \emph{\scriptsize{-0.05}}	\\

\bottomrule
	\end{tabular}
}
\end{table*}


\subsection{Quantitative evaluation}

 \para{\bf Evaluation on the DAVIS dataset:} We compare the performance of our approach to several baselines using the DAVIS dataset. The results are summarized in \tabref{tab:IoU}, where we report the IoU, the contour accuracy, and the time stability metrics. The best method is emphasized in bold font and the second best is underlined. We observe our approach to be quite competitive, outperforming a wide variety of existing unsupervised video segmentation techniques, \eg, 
NLC~\cite{FaktorBMVC14}, MSG~\cite{BroxECCV2010}, KEY~\cite{LeeICCV11}, FST~\cite{PapazoglouICCV13}, FSG~\cite{jain2017fusionseg}, LMP~\cite{TokmakovCVPR2017}, ARP~\cite{KohCVPR17}. We also evaluate our method in the semi-supervised setting by simply replacing the saliency initialization of the first frame with the ground truth. Note that it is common to refer to usage of the first frame as `semi-supervised.' Our unsupervised version is denoted as {\bf OURS-U} and the semi-supervised version is referred to via {\bf OURS-S} in \tabref{tab:IoU}. Semi-supervised baselines are 
SEA~\cite{AvinashCVPR14}, HVS~\cite{GrundmannCVPR2010}, JMP~\cite{FanTOG15}, FCP~\cite{PerazziICCV15}, BVS~\cite{MaerkiCVPR16}, OFL~\cite{TsaiCVPR2016}, CTN~\cite{JangCVPR17}, VPN~\cite{JampaniCVPR17}, and MSK~\cite{PerazziCVPR2017}. Note that OFL uses deep features, and CTN, VPN, MSK, FSG, and LMP are deep learning based approaches. We observe our method to improve the state-of-the-art performance in IoU metric by $1.3\%$ in the unsupervised setting and by $0.7\%$ in the semi-supervised case. Note that beyond training of edge detectors, no learning is performed in our approach. 

In \tabref{tab:attr}, we compare the average IoU of all DAVIS  sequences, clustered by attributes, \eg, appearance change, dynamic blur, fast motion, motion blur, and occlusion. Our method is more robust and outperforms the baselines for fast motion, motion blur and occlusion. In particular, our method performs well for objects with occlusion, outperforming other methods by 10$\%$ for this attribute.

\begin{table}[t]
	\centering 
	\caption{Performance in IoU on SegTrack v2 dataset~\cite{LiICCV13}. }
	\label{tab:segTrack}
\resizebox{0.7\linewidth}{!}{
\setlength\tabcolsep{3pt}

\begin{tabular}{lccccc}
	\toprule
	Sequence	&	KEY~\cite{LeeICCV11}	&	FST~\cite{PapazoglouICCV13}	&	NLC~\cite{FaktorBMVC14}	&	FSG~\cite{jain2017fusionseg}	&	Ours	\\
\midrule
\textsc{birdfall}	&	0.490	&	0.014	&	{\ul 0.565}	&	0.380	&	\textbf{0.649}	\\
\textsc{bird of paradise}	&	{\ul 0.922}	&	0.837	&	0.814	&	0.699	&	{\bf 0.937}	\\
\textsc{bmx}	&	0.630	&	0.621	&	{\ul 0.754}	&	0.591	&	\textbf{0.847}	\\
\textsc{cheetah}	&	0.281	&	0.396	&	{\ul 0.518}	&	\textbf{0.596}	&	{\ul 0.518}	\\
\textsc{drift}	&	0.469	&	0.811	&	0.741	&	\textbf{ 0.876}	&	{\ul 0.829}	\\
\textsc{frog}	&	0.000	&	0.629	&	{\ul 0.713}	&	0.570	&	\textbf{0.832}	\\
\textsc{girl}	&	\textbf{0.877}	&	0.441	&	{\ul 0.860}	&	0.667	&	0.846	\\
\textsc{hummingbird}	&	0.602	&	0.335	&	{\ul 0.624}	&	{\bf 0.652}	&	{0.464}	\\
\textsc{monkey}	&	0.790	&	0.699	&	\textbf{0.823}	&	{\ul 0.805}	&	0.739	\\
\textsc{monkeydog}	&	0.396	&	{\ul 0.523}	&	\textbf{0.525}	&	0.328	&	0.381	\\
\textsc{parachute}	&	\textbf{0.963}	&	0.839	&	0.859	&	0.516	&	{\ul 0.937}	\\
\textsc{penguin}	&	0.093	&	0.074	&	{ 0.139}	&	\textbf{0.713}	&	{\ul 0.240}	\\
\textsc{soldier}	&	0.666	&	0.453	&	0.692	&	{\ul 0.698}	&	\textbf{0.800}	\\
\textsc{worm}	&	\textbf{0.844}	&	0.705	&	0.782	&	0.506	&	{\ul 0.800}	\\
\midrule
Average IoU	&	0.573	&	0.527	&	{\ul 0.672}	&	0.614	&	\textbf{0.701}	\\
\bottomrule
\end{tabular}
}
\end{table}


\begin{table}[t]
	\centering 
	\caption{Performance in IoU on FBMS-59 test set~\cite{OchsPAMI2014}.}
	\label{tab:FBMS}
\resizebox{0.7\linewidth}{!}{
\setlength\tabcolsep{3pt}
\begin{tabular}{lcccccc}\toprule
 & NLC~\cite{FaktorBMVC14} &  POR~~\cite{ZhangCVPR13}   &  POS~\cite{JangCVPR16}  &FST~\cite{PapazoglouICCV13}   & ARP~\cite{KohCVPR17}   & \bf OURS  \\
	\midrule
Average IoU & 0.445 & 0.473 & 0.542 & 0.555 & {\ul 0.598} & \bf 0.608 \\
	\bottomrule 
\end{tabular}
}
\end{table}


\begin{table}[t]
	\centering
{\small

	\caption{Performance comparisons in IoU on the initialization 
	 on the DAVIS and SegTrack v2 datasets. 
	}
			\label{tab:init_prop_comp}
	\begin{tabular}{@{\extracolsep{7pt}}lccccccccc}
		\toprule
		&\multicolumn{5}{c}{DAVIS} & \multicolumn{4}{c}{Segtrack v2}\\
		\cmidrule{2-6}\cmidrule{7-10}
		 &NLC & FST & FSG & LMP & \bf Ours &NLC & FST & FSG  & \bf Ours\\
Training? & - & - & \checkmark & \checkmark & - & - & - & \checkmark  & - \\
		\cmidrule{2-6}\cmidrule{7-10}
		Initial saliency  &  0.402 & 0.456 & {\bf 0.602} & 0.569 & {\ul 0.575}&  0.419 & 0.389 & {\bf 0.530} & {\ul 0.424}\\
		
		
		
		\bottomrule 
	\end{tabular}	
	}
\end{table}


 \para{\bf Evaluation on the SegTrack v2 dataset:} We assess our approach on the SegTrack v2 dataset using identical choice of parameters.
We show the results in \tabref{tab:segTrack}. 
We observe our method to be competitive on SegTrack v2. Note that the reported performance of NLC differs from~\cite{FaktorBMVC14} as in the evaluation in~\cite{FaktorBMVC14} only a subset of the $12$ video sequences were used. We ran the code released by~\cite{FaktorBMVC14} and report the results on the full SegTrack v2 dataset with 14 video sequences. The results we report here are similar to the ones reported in~\cite{Tokmakov17_2}. 

 \para{\bf Evaluation on the FBMS dataset:} We evaluate our method on the FBMS~\cite{OchsPAMI2014} test set which consists of 22 video sequences. The results are presented in \tabref{tab:FBMS}. We observe our approach to outperform the baselines.

 \para{\bf Comparisons of the saliency estimation:} To illustrate the benefits of the proposed motion saliency estimation, we compare the performance of the proposed initialization with other approaches in~\tabref{tab:init_prop_comp} and observe that the proposed saliency estimation  performs very well. Note that the saliency estimation in our approach is unsupervised as opposed to FSG and LMP which are trained on more than 10,000 images and 2,250 videos, respectively.

\setlength{\figwidth}{0.16\textwidth}

\begin{figure*}[t]
	\begin{center}
		\begin{subfigure}[b]{\figwidth}
			\centering Ground truth \vspace{\figmarginv} \\
			\vspace{-0.1cm}
			\includegraphics[width=\linewidth]{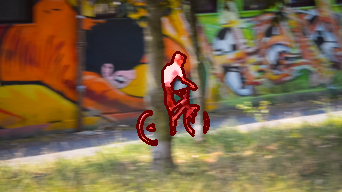} \\
			\includegraphics[width=\linewidth]{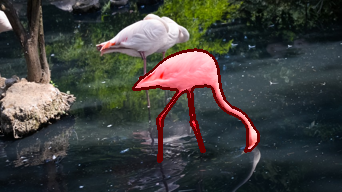} \\
            \includegraphics[width=\linewidth]{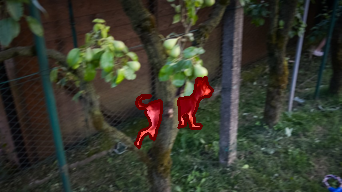} \\
            \includegraphics[width=\linewidth]{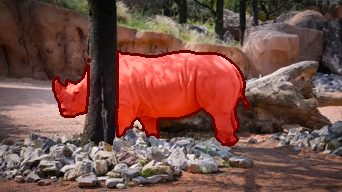} \\
            \includegraphics[width=\linewidth]{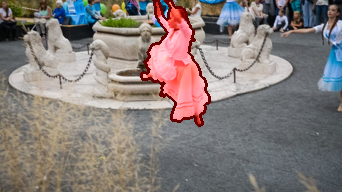} \\
		\end{subfigure}\hfill
		\begin{subfigure}[b]{\figwidth}
			\centering APR \cite{KohCVPR17} \vspace{\figmarginv} \\
			\vspace{-0.1cm}
			\includegraphics[width=\linewidth]{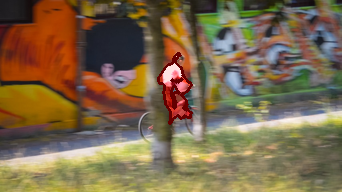} \\
			\includegraphics[width=\linewidth]{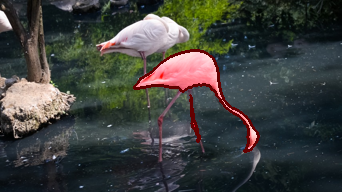} \\
			\includegraphics[width=\linewidth]{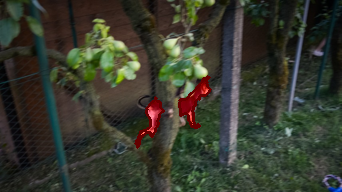} \\
			\includegraphics[width=\linewidth]{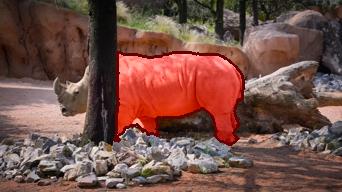} \\
		    \includegraphics[width=\linewidth]{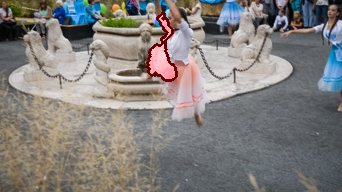} \\
		\end{subfigure}\hfill
		\begin{subfigure}[b]{\figwidth}
			\centering LMP~\cite{TokmakovCVPR2017} \vspace{\figmarginv} \\
			\vspace{-0.1cm}
			\includegraphics[width=\linewidth]{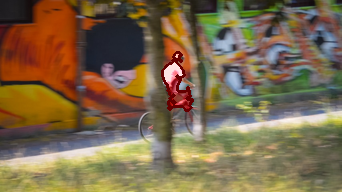} \\
			\includegraphics[width=\linewidth]{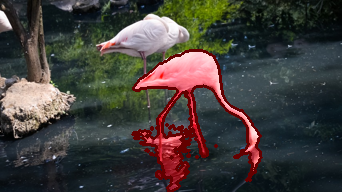} \\
			\includegraphics[width=\linewidth]{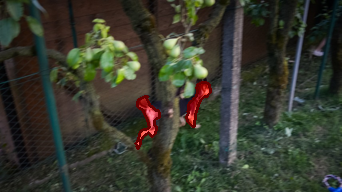} \\
			\includegraphics[width=\linewidth]{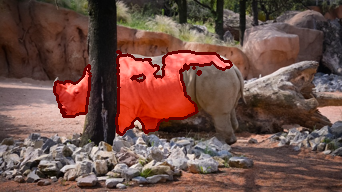} \\
			\includegraphics[width=\linewidth]{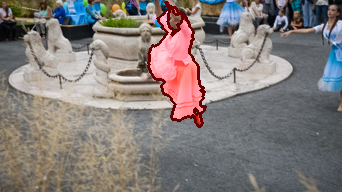} \\
		\end{subfigure}\hfill
		\begin{subfigure}[b]{\figwidth}
			\centering FSG~\cite{jain2017fusionseg} \vspace{\figmarginv} \\
			\vspace{-0.1cm}
			\includegraphics[width=\linewidth]{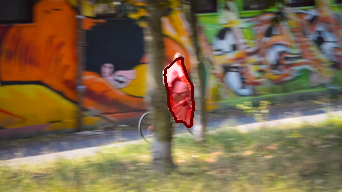} \\
			\includegraphics[width=\linewidth]{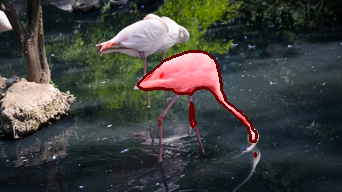} \\
			\includegraphics[width=\linewidth]{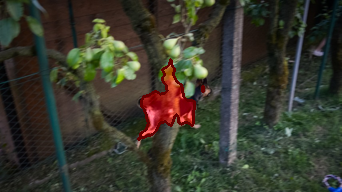} \\
			\includegraphics[width=\linewidth]{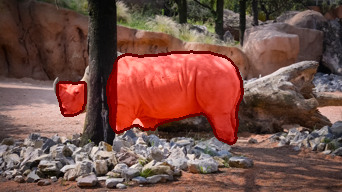} \\
			 \includegraphics[width=\linewidth]{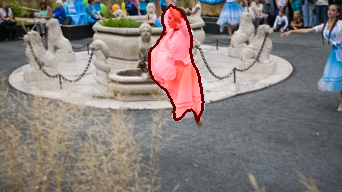} \\
		\end{subfigure}\hfill
		\begin{subfigure}[b]{\figwidth}
			\centering NLC~\cite{FaktorBMVC14} \vspace{\figmarginv} \\
			\vspace{-0.1cm}
			\includegraphics[width=\linewidth]{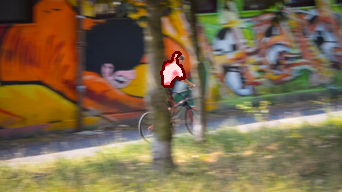} \\
			\includegraphics[width=\linewidth]{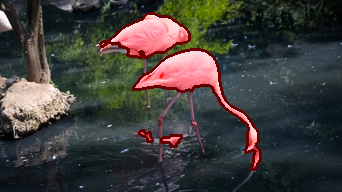} \\
			\includegraphics[width=\linewidth]{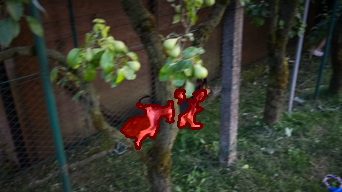} \\
			\includegraphics[width=\linewidth]{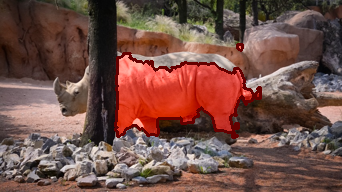} \\
		 \includegraphics[width=\linewidth]{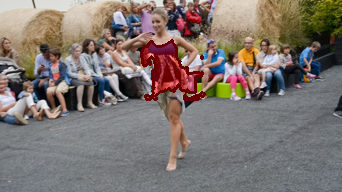} \\
		\end{subfigure}\hfill
		\begin{subfigure}[b]{\figwidth}
			\centering Ours \vspace{\figmarginv} \\
			\vspace{-0.1cm}
			\includegraphics[width=\linewidth]{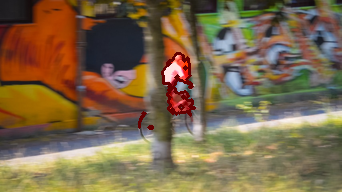} \\
			\includegraphics[width=\linewidth]{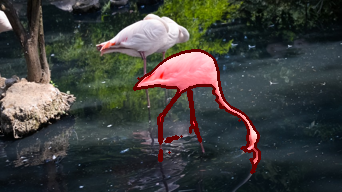} \\
			\includegraphics[width=\linewidth]{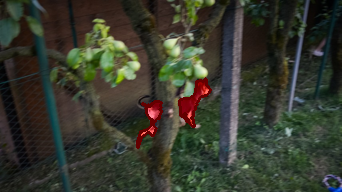} \\
			\includegraphics[width=\linewidth]{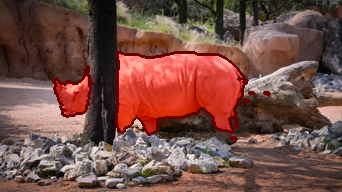} \\
		 \includegraphics[width=\linewidth]{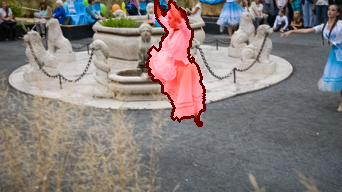} \\
		\end{subfigure}
		\caption{{\tb{Comparison}} of our algorithm and other unsupervised methods on sequence \textsc{bmx-trees} (1st row), \textsc{flamingo} (2nd row), \textsc{libby} (3rd row), \textsc{rhino} (4th row), and \textsc{dance-jump} (5th row) 
		of the DAVIS dataset.}
		\label{fig:comp}
	\end{center}
\end{figure*}

\setlength{\figwidth}{0.16\textwidth}

\begin{figure*}
	\begin{center}
		\begin{subfigure}[b]{\figwidth}
			\includegraphics[width=\linewidth]{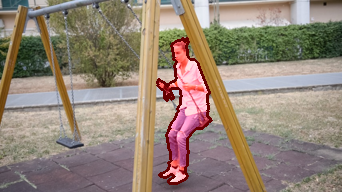} \\
		    \includegraphics[width=\linewidth]{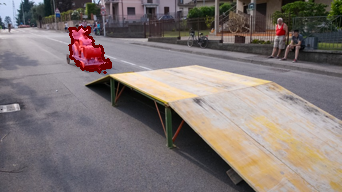} \\
			\includegraphics[width=\linewidth]{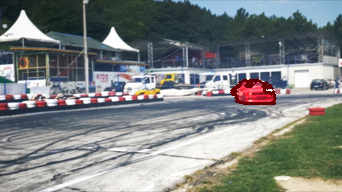} \\
			\includegraphics[width=\linewidth]{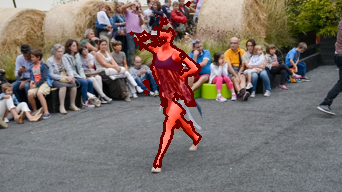} \\
		\end{subfigure}\hfill
		\begin{subfigure}[b]{\figwidth}
			\includegraphics[width=\linewidth]{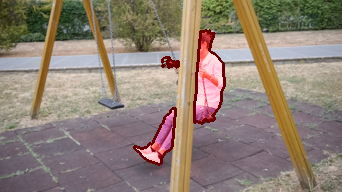} \\
			\includegraphics[width=\linewidth]{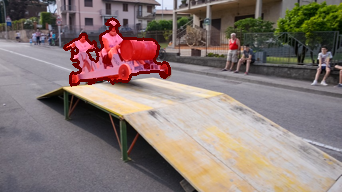} \\
			\includegraphics[width=\linewidth]{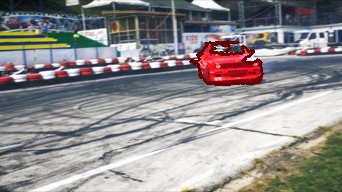} \\
			\includegraphics[width=\linewidth]{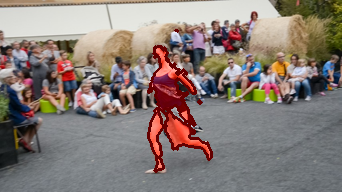} \\
		\end{subfigure}\hfill
		\begin{subfigure}[b]{\figwidth}
			\includegraphics[width=\linewidth]{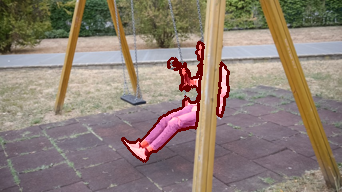} \\
			\includegraphics[width=\linewidth]{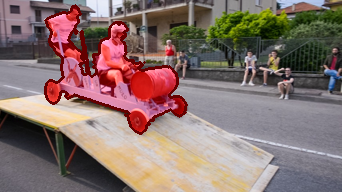} \\
			\includegraphics[width=\linewidth]{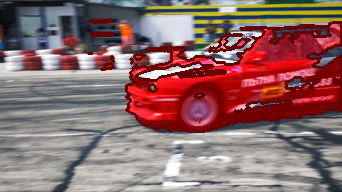} \\
			\includegraphics[width=\linewidth]{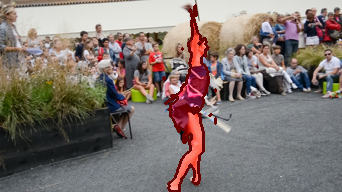} \\
		\end{subfigure}\hfill
		\begin{subfigure}[b]{\figwidth}
			\includegraphics[width=\linewidth]{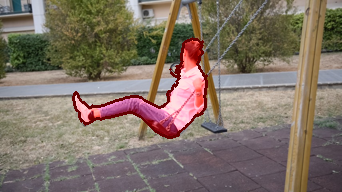} \\
			\includegraphics[width=\linewidth]{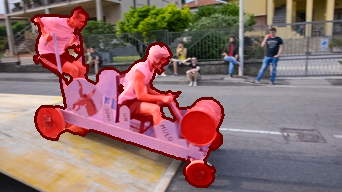} \\
			\includegraphics[width=\linewidth]{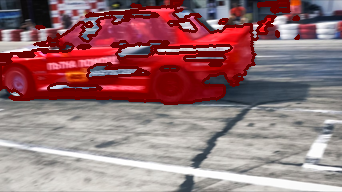} \\
			\includegraphics[width=\linewidth]{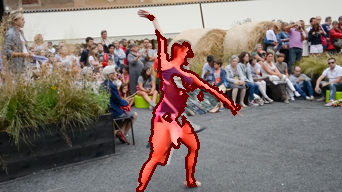} \\
		\end{subfigure}\hfill
		\begin{subfigure}[b]{\figwidth}
			\includegraphics[width=\linewidth]{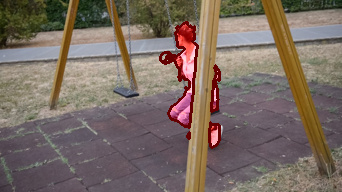} \\
			\includegraphics[width=\linewidth]{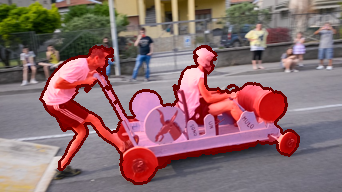} \\
			\includegraphics[width=\linewidth]{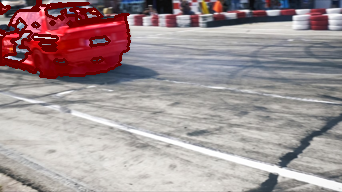} \\
			\includegraphics[width=\linewidth]{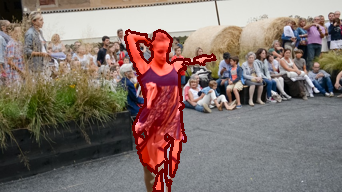} \\
		\end{subfigure}\hfill
		\begin{subfigure}[b]{\figwidth}
			\includegraphics[width=\linewidth]{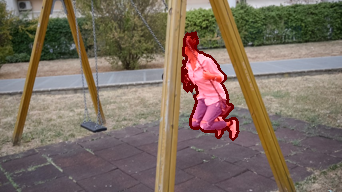} \\
			\includegraphics[width=\linewidth]{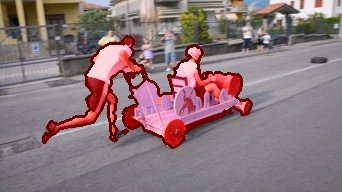} \\
			\includegraphics[width=\linewidth]{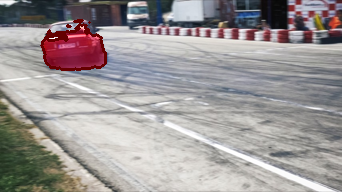} \\
			\includegraphics[width=\linewidth]{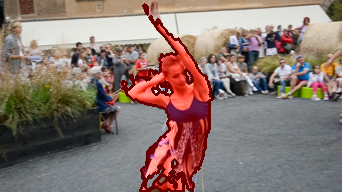} \\
		\end{subfigure}
		\caption{\tb{Visual results} of our approach on the sequences \textsc{swing} (1st row), \textsc{soapbox} (2nd row), \textsc{drift-straight} (3rd row), and \textsc{dance-twirl} (4th row) 
		of the DAVIS dataset.	}
		\label{fig:success}
	\end{center}
\end{figure*}


\begin{figure}[t]
	\begin{center}
		\begin{subfigure}[b]{1.48\figwidth}%
			\includegraphics[width=\linewidth]{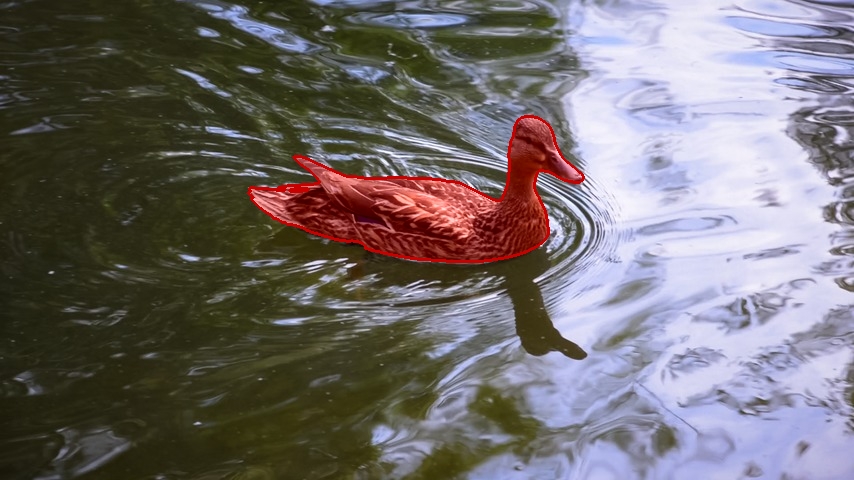}
		\end{subfigure}
		\begin{subfigure}[b]{1.48\figwidth}%
			\includegraphics[width=\linewidth]{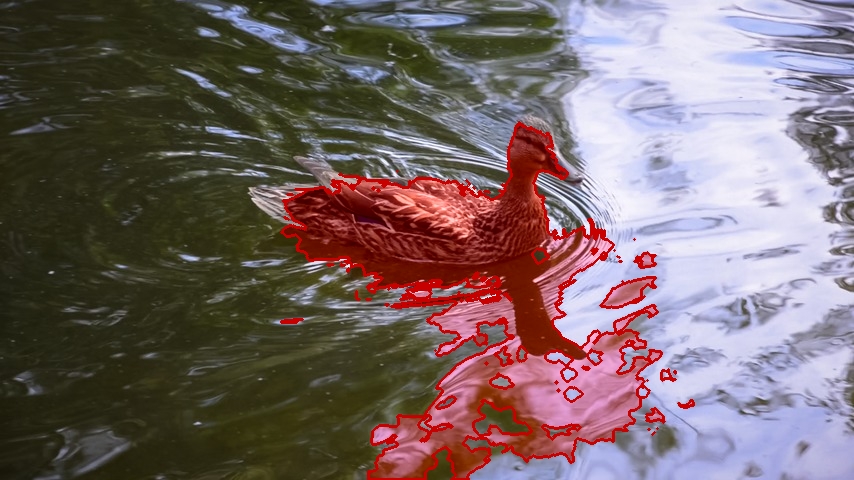}
		\end{subfigure}\hfill
	\end{center}
	\caption{\tb{Failure case.} Groundtruth vs. our result. 
	}
	\label{fig:fail}
\end{figure}

\subsection{Qualitative evaluation}
\para{\bf Side-by-side comparison:} Next we present qualitative results comparing our algorithm to competing methods on challenging parts of the DAVIS dataset. In \figref{fig:comp} we provide side-by-side comparisons to existing methods, \ie, 
APR~\cite{KohCVPR17}, LMP~\cite{TokmakovCVPR2017}, FSG~\cite{jain2017fusionseg}, and NLC~\cite{FaktorBMVC14}. We observe our approach to yield encouraging results even in challenging situations such as  frames in \textsc{bmx-trees} (\figref{fig:comp}, first row), where the foreground object is very small and occluded, and the background is very colorful, and in \textsc{flamingo} (\figref{fig:comp}, second row), where there is non-rigid deformation, and the background object is similar to the foreground object.
We refer the interested reader to the supplementary material for additional results and videos. 

\para{\bf Success cases:}
In \figref{fig:success}, we provide success cases of our algorithm, \ie, frames where our designed technique delineates the foreground object accurately. We want to highlight that our approach is more robust to challenges such as occlusions, motion blur and fast moving objects as the attribute-based aggregate performance in~\tabref{tab:attr} suggests.

\para{\bf Failure modes:}
In \figref{fig:fail}, we also present failure modes of our approach. We observe our technique to be challenged by complex motion. 
Since our method mainly relies on motion and appearance, water is classified as foreground due to its complex motion (\textsc{mallard-water}). 
\section{Conclusion}
We proposed a saliency estimation and a graph neighborhood for effective unsupervised foreground-background video segmentation. Our key novelty is a motion saliency estimation and an informative neighborhood structure. 
Our unsupervised method demonstrates how to effectively exploit the structure of video data, \ie, taking advantage of flow and edges, and achieves state-of-the-art performance in the unsupervised setting.

\noindent\textbf{Acknowledgments:} This material is based upon work supported in part by the National Science Foundation under Grant No.~1718221, 1755785, Samsung, and 3M. We thank NVIDIA for providing the GPUs used for this research.

\clearpage

\bibliographystyle{splncs04}
\bibliography{egbib,alex}
\end{document}